\documentclass[10pt,twocolumn,letterpaper]{article}

\usepackage[pagenumbers]{cvpr} 
\usepackage{multirow}   
\usepackage{stfloats}

\usepackage{cuted} 

\usepackage{currfile} 

\usepackage{caption} 

\usepackage{xspace}

\usepackage{float}

\definecolor{cvprblue}{rgb}{0.21,0.49,0.74}
\usepackage[pagebackref,breaklinks,colorlinks,allcolors=cvprblue]{hyperref}

\def\confName{CVPR}
\def\confYear{2027}

\title{Gen3DOrient: Generative Modeling of 3D-Structured Latents for Object Orientation}

\author{%
  Zhiyuan Zhang\textsuperscript{1 \textdagger}\qquad
  Guofeng Zhang\textsuperscript{1}\qquad
  Jieneng Chen\textsuperscript{2}\qquad
  Adam Kortylewski\textsuperscript{3}\qquad \\
  [4pt]
  Alan Yuille\textsuperscript{1}\qquad
  Xingrui Wang\textsuperscript{1\textdaggerdbl}\\[4pt]
  {\normalsize
    \textsuperscript{1}Johns Hopkins University\qquad
    \textsuperscript{2}Stanford University\qquad
    \textsuperscript{3}University of Freiburg, Germany}%
}

\begin{document}
\maketitle
\begin{strip}
\vspace*{-30pt}
\centering

\includegraphics[width=\linewidth]{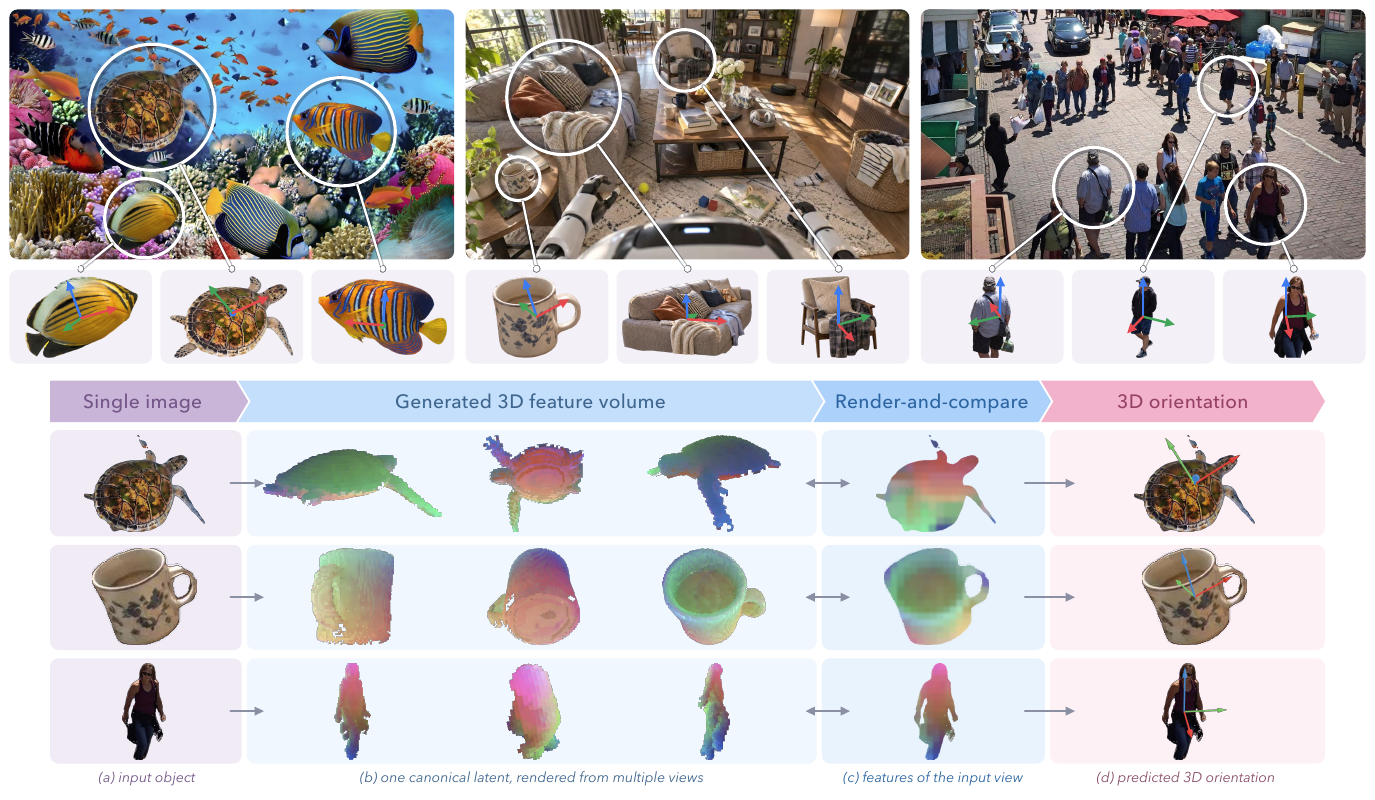}\hfill
\captionsetup{hypcap=false}
\captionof{figure}{Gen3DOrient: generative analysis-by-synthesis for object orientation. From a single image, we generate a canonically aligned, 3D feature volume and infer orientation through feature-space render-and-compare optimization. This enables zero-shot orientation estimation across diverse objects and visual domains without requiring pre-provided object-specific 3D models.}

\captionsetup{hypcap=true}
\label{fig:\currfilebase}%
\end{strip}

\begingroup
  \renewcommand{\thefootnote}{\fnsymbol{footnote}}
  \footnotetext[2]{Email (Zhiyuan Zhang): yx11wng@163.com.}
  \footnotetext[3]{Corresponding author.}
\endgroup

\begin{abstract}

Analysis-by-synthesis has long been a fundamental approach to 3D perception, enabling geometric reasoning by rendering and comparing hypotheses against visual observations. However, its reliance on explicit 3D representations and rendering assumptions has limited its applicability to diverse, unseen objects. Recent advances in visual foundation models and generative modeling offer an opportunity to revisit this classical paradigm. In this paper, we propose Gen3DOrient, a generative approach to 3D object parsing through canonically aligned, 3D-structured latents. Instead of directly predicting geometric annotations, Gen3DOrient learns canonical 3D feature volumes from multi-view images through differentiable feature rendering supervision. Then we train an image-conditioned flow matching model to generate these representations from a single image. The resulting latents enable analysis-by-synthesis in feature space, where object orientations are inferred through render-and-compare optimization. We evaluate Gen3DOrient on zero-shot absolute orientation and relative rotation estimation across diverse out-of-distribution benchmarks, demonstrating strong generalization to unseen objects and visual domains.
\end{abstract}

\section{Introduction}
\label{sec:intro}

Understanding 3D objects from 2D images has long been a fundamental challenge in computer vision. A classical approach is \textbf{analysis-by-synthesis}, which infers 3D properties by rendering hypotheses and comparing them against visual observations~\cite{wang2021nemo,wen2024foundationpose,li2018deepim}. While conceptually appealing, conventional approaches often rely on known object-specific 3D models and explicit rendering assumptions, limiting their applicability to diverse, unseen objects. This raises a fundamental question: Can we revisit analysis-by-synthesis with modern generative models to enable generalizable 3D perception?

Recent advances in visual foundation models and generative 3D modeling provide a promising opportunity. Visual foundation models offer rich perceptual representations, while generative models learn powerful priors over 3D objects. However, connecting these two capabilities with analysis-by-synthesis requires a suitable intermediate representation: one that preserves \textbf{explicit 3D structure, maintains semantic correspondence across viewpoints, and supports feature-space rendering and matching}. Unlike conventional 3D generation, which primarily targets geometry and appearance synthesis, our goal is to generate 3D representations that can directly support geometric inference.

In this work, we introduce \textbf{Gen3DOrient}, which bridges generative 3D representation learning and classical analysis-by-synthesis through \textbf{canonically aligned, 3D-structured latents}. We first construct renderable 3D feature volumes by lifting multi-view visual foundation features onto canonical surface voxels, with feature alignment learned through differentiable rendering supervision. We then train an image-conditioned flow matching model to generate these structured representations from a single image, eliminating the need for an object-specific 3D model at inference time. Finally, we perform analysis-by-synthesis in feature space: candidate orientations are evaluated by rendering the generated latents and comparing them against observed image features, with the orientation iteratively optimized to maximize feature agreement. Rather than directly regressing orientation parameters, our approach uses a generated 3D representation to support explicit hypothesis testing.

We evaluate Gen3DOrient on zero-shot absolute orientation and relative rotation estimation across ten diverse benchmarks, demonstrating strong out-of-distribution generalization across object categories, visual domains, and viewpoint changes. We further explore rotational symmetry recognition through the feature-matching landscape, illustrating how the same representation can support additional geometric reasoning. Together, these results suggest a broader direction for generative 3D modeling: moving beyond synthesizing 3D content toward generating structured representations for visual reasoning.

Our main contributions are summarized as follows:
\begin{itemize}
\item \textbf{Generative formulation for object orientation}. We introduce Gen3DOrient, which connects pretrained 3D generative modeling with classical analysis-by-synthesis, enabling zero-shot orientation inference without a pre-provided object-specific 3D model.

\item \textbf{Orientation-oriented 3D latent modeling}. We construct canonically aligned 3D feature volumes from multi-view visual foundation model features and learn to generate these representations from a single image using conditional flow matching.

\item \textbf{Feature-space inference and generalization}. We perform orientation inference through render-and-compare optimization over generated 3D representations, demonstrating strong zero-shot generalization across absolute orientation and relative rotation benchmarks.
\end{itemize}

\section{Related Work}

\subsection{3D Object Orientation and Pose Estimation}

Estimating object orientation from images has been studied
through both direct prediction and correspondence-based
approaches. Recent feed-forward methods, such as Orient
AnythingV2~\cite{wang2026orient}, leverage large-scale
pretraining and geometry-aware visual backbones to predict
absolute orientations, relative rotations, and rotational
symmetries across diverse object categories.

A related line of work addresses 6D object pose estimation.
Methods such as Gen6D~\cite{liu2022gen6d} and
OnePose++~\cite{he2022onepose++} generalize to novel objects
using posed reference observations, while
FoundationPose~\cite{wen2024foundationpose} supports both
CAD-model-based and reference-image-based inference.
These approaches demonstrate strong generalization under
different assumptions about object information available
at test time. In contrast, our method generates a 3D
representation from the input image and estimates its
orientation through feature-space optimization, without
requiring a pre-provided object-specific 3D model.

\subsection{Visual Foundation Models and 3D Generation}

Visual foundation models have enabled transferable
representations for a wide range of perception tasks.
DINOv2~\cite{oquab2023dinov2} learns dense visual features through
self-supervised pretraining, while VGGT~\cite{wang2025vggt}
incorporates multi-view geometric supervision to predict
camera parameters, depth, and 3D point maps. These models
provide useful feature representations for reasoning about
objects across different viewpoints.

In parallel, recent 3D generative models have advanced
single-image 3D reconstruction and generation.
TRELLIS~\cite{xiang2025structured} introduces structured 3D latents
that combine sparse spatial structure with visual features,
enabling image-conditioned generation through rectified
flow-matching models. Building on these advances, we use
generative modeling to produce canonical 3D feature volumes
for orientation inference, rather than treating 3D asset
generation as the final objective.

\subsection{Analysis-by-Synthesis}

Analysis-by-synthesis estimates underlying scene properties
by rendering candidate hypotheses and comparing them with
visual observations. This principle has been explored in
object pose estimation through render-and-compare methods.
For example, DeepIM~\cite{li2018deepim} iteratively refines
object poses by matching rendered images against observed
images, while FoundationPose~\cite{wen2024foundationpose} employs
rendering-based pose refinement and hypothesis selection
for novel objects.

Our work follows the same analysis-by-synthesis principle
but combines it with image-conditioned generation of
3D-structured latents. Specifically, we generate a
canonical 3D feature volume from a single image and
optimize object orientation by comparing rendered features
with the observed image features. This connects generative
3D modeling with feature-space pose optimization without
requiring a pre-existing 3D model of the query object.

\begin{figure*}[t]
    \centering
    \includegraphics[width=\textwidth]{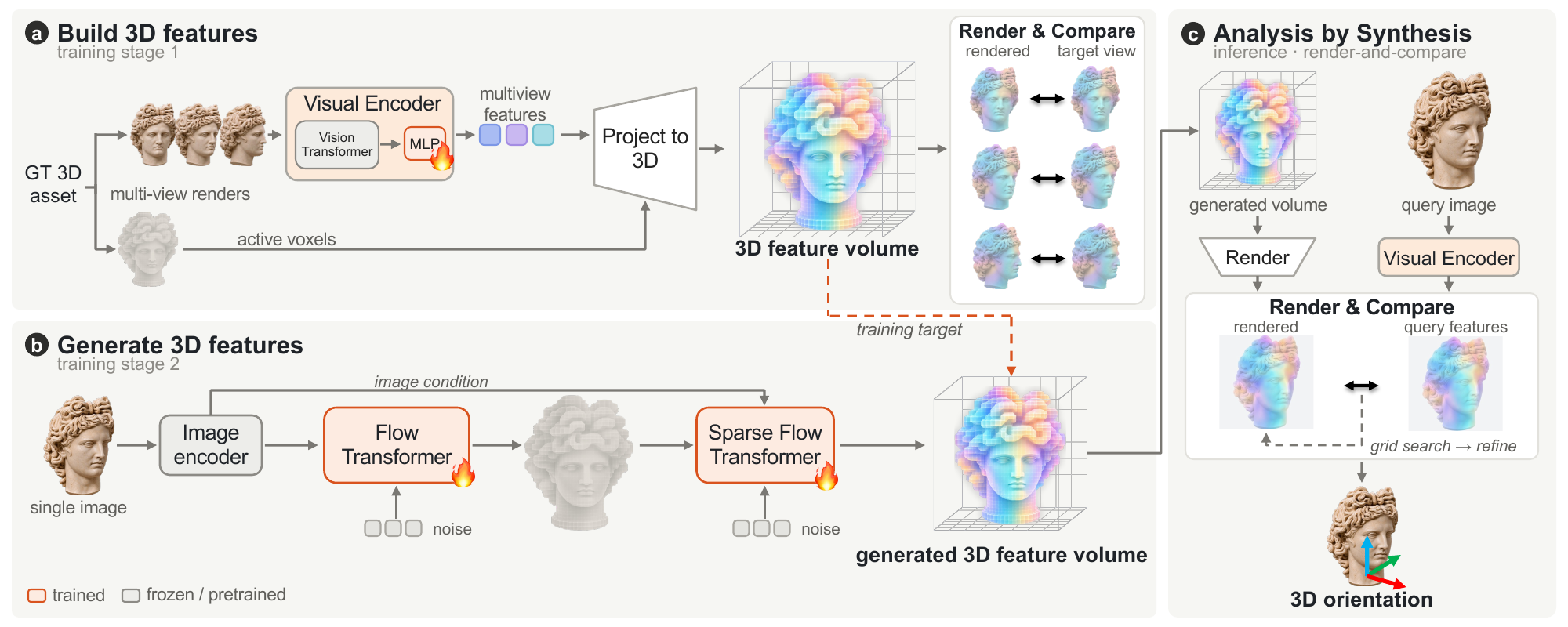}
    \caption{Overview of \textbf{Gen3DOrient}. (a) Canonical 3D feature
    volumes are constructed from aligned multi-view observations and trained
    through feature-rendering supervision. (b) An image-conditioned flow model
    generates the structured representation from a single image.
    (c) Candidate orientations are optimized by comparing rendered latent
    features with observed image features.}
    \label{fig:method}
\end{figure*}

\section{Method}
\label{sec:method}

\subsection{Problem Formulation and Overview}
\label{sec:method_overview}

Given an RGB image $I$ containing an object, our goal is to estimate its
orientation $R \in \mathrm{SO}(3)$ with respect to a canonical object frame.
Unlike template-based pose estimators, we do not assume access to an
object-specific 3D model at test time. We approach this problem by combining
single-image 3D generation with feature-space analysis-by-synthesis.
Specifically, an image-conditioned generative model $G_{\theta}$ produces a
canonical, 3D-structured latent $\hat{\mathcal V}=G_{\theta}(I,z)$ from noise
$z$. Given a feature encoder $E_{\psi}$ and a pose-conditioned feature
renderer $\mathcal R$, we then estimate orientation by
\begin{equation}
\label{eq:overview}
\hat R = \arg\min_{R\in\mathrm{SO}(3)}
\mathcal D\!\left(\mathcal R(\hat{\mathcal V},R),\,E_{\psi}(I)\right),
\end{equation}
where $\mathcal D$ measures agreement between rendered and observed feature
maps. The generated latent and visual encoder remain fixed during this
optimization; only the orientation is updated. For relative rotation, we
estimate the orientations of the query and reference images in the shared
canonical frame and compose the two rotations.

Our framework consists of three parts (Fig.~\ref{fig:method}): construction
of canonical 3D feature targets from multi-view training data
(Sec.~\ref{sec:method_latents}), image-conditioned generation of those targets
(Sec.~\ref{sec:method_generation}), and orientation inference through
feature-space render-and-compare (Sec.~\ref{sec:method_inference}).

\subsection{Canonical 3D-Structured Latents}
\label{sec:method_latents}

\textbf{Canonical sparse volume.}
Our generative target is a sparse 3D feature volume, rather than a textured
mesh or an unstructured image embedding. During training, we assume an
object mesh $\mathcal M$ with a consistent canonical orientation. We
voxelize its surface on a $64^3$ grid and record the set of occupied voxel
coordinates $\mathcal S=\{\mathbf p_k\}_{k=1}^{K}$. Each occupied voxel is
assigned a $D$-dimensional feature vector $\mathbf f_k\in\mathbb R^D$
($D=1024$ in our implementation). We denote the resulting representation by
\begin{equation}
    \mathcal V=(\mathcal S,\mathcal F),\qquad
    \mathcal F=\{\mathbf f_k\}_{k=1}^{K}.
\end{equation}
The geometry specifies where features reside in the canonical object frame;
the feature vectors are obtained from visual observations. Canonical pose
calibration and mesh preparation are described in Sec.~\ref{sec:experiments}.

\textbf{Multi-view feature lifting.}
For each training object, we render images $\{I_j\}_{j=1}^{N}$ with known
cameras $\{P_j\}_{j=1}^{N}$. Our visual encoder consists of pretrained
DINOv2\cite{oquab2023dinov2} and VGGT components~\cite{wang2025vggt}, followed by a lightweight
alignment MLP $A_{\psi}$. The pretrained components are frozen; only
$A_{\psi}$ is trained. We write $E_{\psi}=A_{\psi}\circ E_0$ for the resulting
feature extractor, with $E_0$ denoting the frozen visual backbone.

We lift the 2D features of source views onto the canonical surface voxels
using the known cameras and visibility-aware projection. At voxel $k$, the
features sampled from its visible projections are aggregated into
$\mathbf f_k$ using visibility weights. This produces a 3D volume whose
features can be rendered from a different viewpoint without changing its
canonical coordinate frame.

\textbf{Feature-rendering supervision.}
To learn the alignment MLP, we reserve a target view $i$ and construct the
volume from the remaining source views. We render this volume at the known
target camera $P_i$ to obtain $\hat F_i=\mathcal R(\mathcal V,P_i)$, and
compare it with the target-view feature map $F_i^{\star}=E_{\psi}(I_i)$.
With $w_i(u)$ denoting a coverage-derived weight at pixel $u$, the
alignment objective is
\begin{equation}
\label{eq:align}
\mathcal L_{\mathrm{align}}=
\frac{\sum_{i}\sum_{u}w_i(u)\,
\left\|\hat F_i(u)-F_i^{\star}(u)\right\|_2^2}
{D\sum_{i}\sum_{u}w_i(u)}.
\end{equation}
Both feature maps are channel-wise $\ell_2$-normalized. This supervision
optimizes the alignment module to make lifted multi-view features agree
with features extracted from held-out viewpoints. After training, we
construct and save a canonical volume for each training object as the
target of the generative model.

\subsection{Generative Modeling of 3D-Structured Latents}
\label{sec:method_generation}

At inference time, the object-specific mesh used to construct
$\mathcal V$ is unavailable. We therefore learn an image-conditioned
generative model that predicts both the sparse 3D structure and its
associated feature vectors from a single image. We adopt the pretrained
TRELLIS generative framework~\cite{xiang2025structured} as the initialization and
architectural starting point, while changing the feature-generation
target to the canonical volumes constructed above.

\textbf{Sparse structure and feature generation.}
Following the two-part organization of structured 3D generation, a
structure model first predicts the occupied voxel coordinates
$\hat{\mathcal S}$ from the input image. Conditioned on this structure and
on the image, a feature model then generates the vectors
$\hat{\mathcal F}$ attached to active voxels. The final output is
$\hat{\mathcal V}=(\hat{\mathcal S},\hat{\mathcal F})$.

The key distinction from TRELLIS's conventional asset-generation pipeline
lies in the feature target and its use: our local feature generator
predicts $D$-dimensional, foundation-feature-aligned descriptors used
directly by the pose renderer, rather than the original compact latent
codes intended for appearance decoding. Accordingly, in the local
feature-generation branch we retain the transformer's DiT core and remove
the original sparse-convolutional compression and upsampling modules that
map to the compact output codes. We do not use an appearance VAE to decode
these generated descriptors into RGB textures. The sparse-structure
branch is treated separately and follows the adopted structure-generation
pipeline.

\textbf{Conditional flow-matching objective.}
Both stages use image-conditioned flow matching. Let $x_0$ denote a clean
training target for a given stage (a structure representation or local
features), let $\epsilon\sim\mathcal N(0,\mathbf I)$, and let
$t\sim\mathcal U[0,1]$. We form
\begin{equation}
    x_t=(1-t)x_0+t\epsilon,
\end{equation}
and train a velocity predictor $v_{\theta}$ using
\begin{equation}
\label{eq:cfm}
\mathcal L_{\mathrm{FM}}=
\mathbb E_{x_0,\epsilon,t,I}
\left[\left\|v_{\theta}(x_t,t,c(I))
-(\epsilon-x_0)\right\|_2^2\right],
\end{equation}
where $c(I)$ is the image condition; for local feature generation, the
predicted sparse structure is also provided as conditioning. We train the
structure and feature models for their respective targets. At inference,
we sample the structure followed by its feature vectors, yielding a
canonical 3D feature volume for render-and-compare optimization.

\begin{table*}[t]
\centering
\setlength{\tabcolsep}{5.1pt}
\renewcommand{\arraystretch}{1.08}
\resizebox{\textwidth}{!}{%
\begin{tabular}{l*{11}{c}}
\toprule
\multirow{2}{*}{Method}
 & \multicolumn{2}{c}{SUN-RGBD}
 & \multicolumn{2}{c}{ARKitScenes}
 & \multicolumn{2}{c}{Pascal3D+}
 & \multicolumn{2}{c}{Objectron}
 & \multicolumn{2}{c}{ImageNet3D}
 & Ori\_COCO \\
\cmidrule(lr){2-3}\cmidrule(lr){4-5}\cmidrule(lr){6-7}
\cmidrule(lr){8-9}\cmidrule(lr){10-11}\cmidrule(lr){12-12}
 & Med$\downarrow$ & Acc30$\uparrow$
 & Med$\downarrow$ & Acc30$\uparrow$
 & Med$\downarrow$ & Acc30$\uparrow$
 & Med$\downarrow$ & Acc30$\uparrow$
 & Med$\downarrow$ & Acc30$\uparrow$
 & Acc$\uparrow$ \\
\midrule
OrientAnythingV2 & 26.00 & 55.4 & 36.48 & 43.2 & 15.02 & 72.7 & 22.62 & 56.4 & 15.26 & 65.2 & 86.4 \\
SAM 3D           & 34.14 & 45.7 & 53.80 & 35.4 & 28.16 & 63.2 & 48.28 & 40.0 & 29.03 & 51.7 & 60.2 \\
\midrule
\textbf{Gen3DOrient} & \textbf{23.87} & \textbf{59.1} & \textbf{32.84} & \textbf{45.7}
& \textbf{13.46} & \textbf{73.8} & \textbf{21.04} & \textbf{57.8}
& \textbf{13.71} & \textbf{67.9} & \textbf{88.4} \\
\bottomrule
\end{tabular}%
}
\caption{\textbf{Zero-shot absolute orientation estimation.}
Median rotation error (Med, degrees; lower is better), accuracy within
$30^\circ$ (Acc30; higher is better), and discrete orientation accuracy
on Ori\_COCO (Acc; higher is better). Best reported results are bold.}
\label{tab:orientation_estimation}
\end{table*}

\begin{table*}[t]
\centering
\setlength{\tabcolsep}{4.2pt}
\renewcommand{\arraystretch}{1.07}
\resizebox{\textwidth}{!}{%
\begin{tabular}{l*{12}{c}}
\toprule
\multirow{2}{*}{Method}
& \multicolumn{3}{c}{LINEMOD}
& \multicolumn{3}{c}{YCB-Video}
& \multicolumn{3}{c}{OnePose++}
& \multicolumn{3}{c}{OnePose} \\
\cmidrule(lr){2-4}\cmidrule(lr){5-7}\cmidrule(lr){8-10}\cmidrule(lr){11-13}
& Med$\downarrow$ & Acc30$\uparrow$ & Acc15$\uparrow$
& Med$\downarrow$ & Acc30$\uparrow$ & Acc15$\uparrow$
& Med$\downarrow$ & Acc30$\uparrow$ & Acc15$\uparrow$
& Med$\downarrow$ & Acc30$\uparrow$ & Acc15$\uparrow$ \\
\midrule
\multicolumn{13}{c}{\emph{POPE sampling (mean relative rotation: $14.85^\circ$)}} \\
\midrule
Gen6D            & 44.86 & 36.4 & 9.6 & 54.48 & 23.2 & 7.7 & 35.43 & 41.1 & 15.8 & 17.78 & 89.3 & 38.9 \\
POPE             & 15.73 & 77.0 & 48.3 & 13.94 & 80.1 & 54.4 & 6.27 & 89.6 & 72.8 & \textbf{2.16} & 96.2 & 91.1 \\
OrientAnythingV2 & 7.82 & 98.07 & 89.7 & \textbf{6.07} & 91.6 & 86.4 & 6.18 & \textbf{99.7} & \textbf{96.6} & 6.76 & 99.7 & 95.7 \\
\midrule
\textbf{Gen3DOrient} & \textbf{6.93} & \textbf{98.45} & \textbf{91.6} & 6.12 & \textbf{91.9} & \textbf{86.5}
& \textbf{6.13} & \textbf{99.7} & 96.5 & 6.68 & \textbf{99.8} & \textbf{96.2} \\
\midrule
\multicolumn{13}{c}{\emph{Random-pair sampling (mean relative rotation: $78.22^\circ$)}} \\
\midrule
POPE & 98.03 & 10.3 & 4.3 & 41.88 & 40.9 & 27.2 & 88.21 & 25.6 & 19.8 & 45.73 & 45.1 & 37.3 \\
OrientAnythingV2 & 28.83 & 51.6 & 28.3 & 15.78 & 61.2 & 48.7 & 12.83 & 85.5 & 58.8 & 11.72 & 86.7 & 63.4 \\
\midrule
\textbf{Gen3DOrient} & \textbf{24.72} & \textbf{54.1} & \textbf{32.9}
& \textbf{14.63} & \textbf{61.8} & \textbf{50.1}
& \textbf{8.46} & \textbf{89.3} & \textbf{63.7}
& \textbf{8.05} & \textbf{91.4} & \textbf{68.2} \\
\bottomrule
\end{tabular}%
}
\caption{\textbf{Zero-shot relative rotation estimation.}
Results for reference--query pairs under POPE sampling and random-pair
sampling. Med is the median rotation error in degrees; Acc30 and Acc15
measure accuracy within $30^\circ$ and $15^\circ$, respectively. Bold
indicates the best reported result within each sampling setting;
ties are both bold.}
\label{tab:relative_rotation}
\end{table*}

\begin{table}[t]
\centering
{\footnotesize
\setlength{\tabcolsep}{3pt}
\begin{tabular*}{\columnwidth}{@{\extracolsep{\fill}}lcccc@{}}
\toprule
\multirow{2}{*}{Method}
& \multicolumn{2}{c}{Symmetry Acc$\uparrow$}
& \multicolumn{2}{c}{Animal3D} \\
\cmidrule(lr){2-3}\cmidrule(l){4-5}
& OA-V2 & Extended & Med$\downarrow$ & Acc30$\uparrow$ \\
\midrule
OrientAnythingV2
& 65.2 & 62.7 & 15.28 & 69.0 \\
\textbf{Gen3DOrient}
& \textbf{69.5} & \textbf{68.1} & \textbf{12.96} & \textbf{73.0} \\
\bottomrule
\end{tabular*}\par
}
\caption{\textbf{Zero-shot horizontal symmetry recognition and animal
orientation estimation.}
Symmetry accuracy (\%) is reported on OA-V2 ($N\in\{0,1,2,4\}$)
and the extended set ($N\in\{0,1,2,4,8\}$), where $N$ is the symmetry
order. On Animal3D, Med is the median rotation error in degrees and
Acc30 is the percentage of predictions within $30^\circ$.}
\label{tab:animals_symmetry}
\end{table}

\subsection{Feature-Space Analysis-by-Synthesis}
\label{sec:method_inference}

\textbf{Pose-conditioned feature rendering.}
Given the generated volume $\hat{\mathcal V}$, we represent the exposed
faces of occupied voxels as a surface for rasterization. Each surface
face inherits the feature vector of its voxel. Rendering at a candidate
orientation $R$ yields a 2D feature map
$F_{\mathrm{hard}}(R)=\mathcal R_{\mathrm{hard}}
(\hat{\mathcal V},R)$, in which visible pixels receive the feature of
the frontmost surface element and background pixels receive zero. We
compare this rendered map with the feature map
$F_q=E_{\psi}(I)$ extracted from the query image using the same
coverage-weighted feature distance as Eq.~\eqref{eq:align}.

\textbf{Differentiable orientation refinement.}
The discrete surface assignment in the forward renderer gives weak or
undefined gradients with respect to the orientation. We therefore use a
straight-through rendering formulation. In addition to the hard renderer,
we compute a soft feature map $F_{\mathrm{soft}}(R)$ by projecting voxel
centers to the image and aggregating their features with Gaussian spatial
weights. The renderer used for optimization is
\begin{equation}
\label{eq:straight_through}
F_{\mathrm{ST}}(R)=F_{\mathrm{soft}}(R)
+\operatorname{sg}\!\left(F_{\mathrm{hard}}(R)
-F_{\mathrm{soft}}(R)\right),
\end{equation}
where $\operatorname{sg}(\cdot)$ stops gradients. Thus, forward values
follow the hard surface renderer, while gradients with respect to $R$
are supplied by the soft projection. We hold the generated volume and
feature encoder fixed and optimize only orientation parameters.

\textbf{Initialization and pose selection.}
To reduce sensitivity to local minima, we evaluate a set of candidate
orientations covering azimuth, polar angle, and in-plane rotation.
The candidate with the smallest feature-matching loss initializes
gradient-based refinement with Adam. This produces the final absolute
orientation. For reference-based relative rotation, the same procedure
is applied to the reference and query images, and the resulting rotations
are composed in the shared canonical frame. No object-specific mesh or
per-object test-time model fitting is required.

\begin{figure*}[t]
    \centering
    \includegraphics[width=\textwidth]{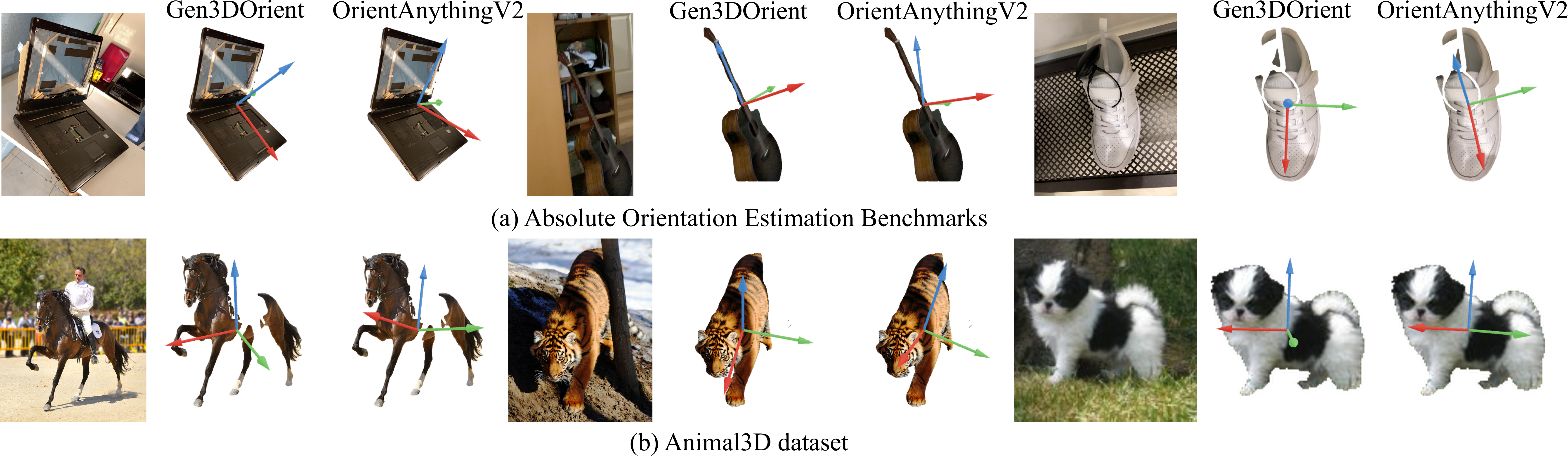}
    \caption{\textbf{Qualitative results of zero-shot absolute orientation estimation.}
    The first row is selected objects from the absolute orientation estimation benchmarks and the second row is selected objects from animal dataset. For each object, the panels display (from left to right): the input RGB image, and the estimated orientations from Gen3DOrient (via render-and-compare) and OrientAnythingV2.}
    \label{fig:orientation_qualitative}
\end{figure*}

\subsection{Data Curation}
\label{sec:data_curation}

Our training data are drawn from the OrientAnythingV2\cite{wang2026orient} and ImageNet3D\cite{ma2024imagenet3d}
collections, reported to contain approximately 600K and 70K objects,
respectively. To construct canonical training targets, we use
OrientAnythingV2~\cite{wang2026orient} to calibrate object orientations,
followed by VLM-assisted\cite{bai2025qwen3} filtering and correction of inconsistent labels.
For each 3D object, we uniformly sample 16 views, extract multi-view
features, and lift them onto surface voxels of the canonicalized mesh.
After learning feature alignment, we precompute the resulting 3D feature
volumes as targets for image-conditioned generation, as described in
Sec.~\ref{sec:method_latents}.

\section{Experiments}
\label{sec:experiments}
\vspace{-0.5em}

We evaluate Gen3DOrient primarily through zero-shot object orientation
estimation. Our experiments address three questions: (i) how well does the
model transfer across datasets for absolute orientation estimation,
(ii) whether the same framework generalizes to reference-based relative
rotation, especially under large viewpoint changes, and (iii) which
components are important for its performance. We additionally examine
rotational symmetry recognition as an application of the feature-matching
objective beyond estimating a single optimal orientation.

\subsection{Experimental Setup}
\label{sec:exp_setup}

\textbf{Training configuration.}
We optimize the feature-alignment stage with AdamW for 30K iterations,
using a batch size of 224 on eight GPUs. For the generative stage, the
structure flow transformer is trained for 50K iterations with a batch
size of 32, and the sparse feature flow transformer is trained for 50K iterations
with a batch size of 8, also on eight GPUs. Both use AdamW optimizer and a learning rate of $1\times10^{-4}$. Further
optimizer settings and inference-time hyperparameters are provided in
the supplementary material.

\textbf{Evaluation protocol.}
We report results without fine-tuning on the target benchmarks. For
absolute and relative rotation, the median rotation error
(\textbf{Med}, lower is better) is computed in degrees. We additionally
report the percentage of predictions within $30^\circ$ (\textbf{Acc30})
and, where available, $15^\circ$ (\textbf{Acc15}). For the discrete
horizontal-orientation benchmark Ori\_COCO, we use classification
accuracy (\textbf{Acc}).

\subsection{Zero-Shot Absolute Orientation Estimation}
\label{sec:exp_absolute}

\textbf{Benchmarks and baselines.}
We evaluate on SUN-RGBD\cite{song2015sun}, ARKitScenes\cite{baruch2021arkitscenes}, Pascal3D+\cite{xiang2014beyond}, Objectron\cite{ahmadyan2021objectron}, ImageNet3D\cite{ma2024imagenet3d},
and Ori\_COCO\cite{wang2024orient}. For ImageNet3D, we evaluate exclusively on its held-out test split, which is strictly disjoint from the training subset described in Sec.\ref{sec:data_curation}. The first five benchmarks use Med and Acc30; Ori\_COCO
contains eight horizontal-orientation labels and uses classification
accuracy. We compare with OrientAnythingV2~\cite{wang2026orient} and
SAM 3D~\cite{chen2026sam}, using the reported evaluation protocol.

\textbf{Results.}
Table~\ref{tab:orientation_estimation} shows that Gen3DOrient achieves
the best reported results among the listed methods on all six
benchmarks. Relative to OrientAnythingV2, the Acc30 gain ranges from
1.1 to 3.7 percentage points across the five continuous-orientation
datasets, while median angular error decreases in every case. The
improvements on SUN-RGBD (59.1\% vs. 55.4\% Acc30) and ARKitScenes
(45.7\% vs. 43.2\%) are notable because these benchmarks contain
real-world indoor imagery. On Ori\_COCO, Gen3DOrient improves discrete
orientation accuracy from 86.4\% to 88.4\%.

\begin{table*}[t]
\centering
\footnotesize
\setlength{\tabcolsep}{3pt}
\renewcommand{\arraystretch}{1.08}
\begin{tabular*}{\textwidth}{@{\extracolsep{\fill}}l*{12}{c}@{}}
\toprule
\multirow{2}{*}{Configuration}
& \multicolumn{2}{c}{SUN-RGBD}
& \multicolumn{2}{c}{ARKitScenes}
& \multicolumn{2}{c}{Pascal3D+}
& \multicolumn{2}{c}{Objectron}
& \multicolumn{2}{c}{ImageNet3D}
& Ori\_COCO & Time \\
\cmidrule(lr){2-3}\cmidrule(lr){4-5}
\cmidrule(lr){6-7}\cmidrule(lr){8-9}
\cmidrule(lr){10-11}\cmidrule(lr){12-12}\cmidrule(l){13-13}
& Med$\downarrow$ & Acc30$\uparrow$
& Med$\downarrow$ & Acc30$\uparrow$
& Med$\downarrow$ & Acc30$\uparrow$
& Med$\downarrow$ & Acc30$\uparrow$
& Med$\downarrow$ & Acc30$\uparrow$
& Acc$\uparrow$ & (s) \\
\midrule
Without VLM curation
& 26.93 & 55.8 & 37.13 & 42.5 & 15.21 & 72.5
& 23.17 & 56.1 & 15.73 & 65.9 & 86.0 & -- \\
\addlinespace[2pt]
Grid $12{\times}8{\times}8$
& 30.54 & 45.7 & 42.63 & 38.1 & 17.15 & 69.9
& 24.57 & 53.1 & 17.90 & 63.8 & 83.9 & 9.3 \\
Grid $48{\times}16{\times}16$
& 27.92 & 53.5 & 38.36 & 41.4 & 15.08 & 72.1
& 23.68 & 53.9 & 15.37 & 64.3 & 84.5 & 17.5 \\
\midrule
\textbf{Full configuration}
& \textbf{23.87} & \textbf{59.1}
& \textbf{32.84} & \textbf{45.7}
& \textbf{13.46} & \textbf{73.8}
& \textbf{21.04} & \textbf{57.8}
& \textbf{13.71} & \textbf{67.9}
& \textbf{88.4} & 40.6 \\
\bottomrule
\end{tabular*}
\caption{\textbf{Ablations on training-data curation and pose initialization.}
The full configuration uses VLM-assisted curation and a
$96{\times}32{\times}32$ initialization grid. Each ablation varies one
factor relative to the full configuration. Grid dimensions correspond
to azimuth, polar, and in-plane rotation. The dash indicates an
unreported runtime. Best orientation scores are bold.}
\label{tab:ablation_curation_init}
\end{table*}

\subsection{Zero-Shot Relative Rotation Estimation}
\label{sec:exp_relative}

\textbf{Benchmarks and protocol.}
We evaluate reference-based relative rotation on LINEMOD\cite{hinterstoisser2012model}, YCB-Video\cite{calli2015ycb},
OnePose++\cite{he2022onepose++}, and OnePose\cite{sun2022onepose}. Following the pair-preparation protocol of
POPE~\cite{fan2024pope}, we consider two settings: the original POPE view
pairs, whose average relative rotation is $14.85^\circ$, and randomly
sampled pairs, whose average relative rotation is $78.22^\circ$.
We report Med, Acc30, and Acc15, and compare with the baselines listed
in Table~\ref{tab:relative_rotation}.

\textbf{Results.}
With POPE's original sampling, Gen3DOrient is competitive with the
strongest baselines and achieves the highest Acc30 on LINEMOD,
YCB-Video, and OnePose, while tying OrientAnythingV2 on OnePose++.
Some error and Acc15 metrics remain marginally better for other
methods; we therefore do not claim uniform superiority in this
setting. Under the more challenging random-pair setting, Gen3DOrient
outperforms the listed baselines on all four datasets for all three
reported metrics. Relative to OrientAnythingV2, Acc30 improves by
2.5, 0.6, 3.8, and 4.7 percentage points on LINEMOD, YCB-Video,
OnePose++, and OnePose, respectively. These results support the use of
the same generated representation for orientation inference across
larger viewpoint differences.


\begin{table}[t]
\centering
\setlength{\tabcolsep}{4pt}
\resizebox{\columnwidth}{!}{%
\begin{tabular}{l*{6}{c}}
\toprule
& \multicolumn{2}{c}{Pascal3D+} & \multicolumn{2}{c}{ImageNet3D}
& \multicolumn{2}{c}{VisionEncoderSet} \\
\cmidrule(lr){2-3}\cmidrule(lr){4-5}\cmidrule(lr){6-7}
Encoder & Med$\downarrow$ & Acc30$\uparrow$ & Med$\downarrow$ & Acc30$\uparrow$ & Med$\downarrow$ & Acc30$\uparrow$ \\
\midrule
DINOv2 & 15.59 & 71.3 & 15.68 & 63.7 & 16.42 & 80.9 \\
VGGT & 14.81 & 72.5 & 14.97 & 66.3 & 13.65 & 83.6 \\
VGGT + alignment & \textbf{13.46} & \textbf{73.8} & \textbf{13.71} & \textbf{67.9} & \textbf{11.32} & \textbf{85.2} \\
\bottomrule
\end{tabular}%
}
\caption{\textbf{Ablation on the visual feature encoder.} The
full method uses VGGT features with a learned alignment layer.
}
\label{tab:ablation_3d_encoder}
\end{table}

\subsection{Zero-Shot Rotational Symmetry Recognition}
\label{sec:exp_symmetry}

Rotational symmetry offers an additional test of the information
available in the feature-matching objective. We evaluate horizontal
rotational symmetry orders $N\in\{0,1,2,4\}$ on the OA-V2 benchmark\cite{wang2026orient},
where $N=0$ denotes continuous rotational symmetry and $N=1$
denotes asymmetry. We also use an extended benchmark that includes
20 additionally annotated objects with eight-fold symmetry ($N=8$) from Omni6DPose\cite{zhang2024omni6dpose} dataset.

To infer symmetry using Gen3DOrient, we fix the polar and in-plane angles obtained during
initialization and examine the azimuthal feature-matching loss profile.
A near-flat profile is taken as evidence of continuous symmetry.
Otherwise, we test whether near-minimal loss values recur at evenly
spaced azimuth intervals corresponding to a candidate discrete symmetry
order. We select the largest supported order without training a
symmetry-specific prediction head.

As shown in Table~\ref{tab:animals_symmetry}, Gen3DOrient improves
accuracy from 65.2\% to 69.5\% on OA-V2 benchmark. On the extended benchmark,
which includes $N=8$ cases, it improves accuracy from 62.7\% to 68.1\%. While we adapt OrientAnythingV2 to support $N=8$ by expanding the periodicity search range of its post-processing least-squares fitting before zero-truncation, its pre-trained weights inherently map all $N > 4$ instances to uniform distributions ($N=0$) during training, rendering it largely incapable of disentangling 8-fold rotational symmetry from continuous symmetry at inference.

\subsection{Zero-Shot Orientation Estimation on Animals}

To evaluate the generalization of Gen3DOrient to animals, we manually
annotate the full 3D orientation of one animal instance in each of 100
images from Animal3D~\cite{xu2023animal3d}. Following our zero-shot
evaluation protocol, we report median rotation error (Med) and accuracy
within $30^\circ$ (Acc30). As shown in Table~\ref{tab:animals_symmetry},
Gen3DOrient outperforms OrientAnythingV2, reducing the median error from
$15.28^\circ$ to $12.96^\circ$ and improving Acc30 from 69.0\% to
73.0\%, a gain of 4.0 percentage points.

Figure~\ref{fig:orientation_qualitative} presents qualitative examples for selected
animals, showing the orientation visualization from Gen3DOrient after render-and-compare, and the corresponding prediction from OrientAnythingV2.
Together with the quantitative improvements, these
visualizations support the effectiveness of our generative
analysis-by-synthesis framework for animal orientation estimation.

\begin{figure*}[t]
    \centering
    \includegraphics[width=2.1\columnwidth]{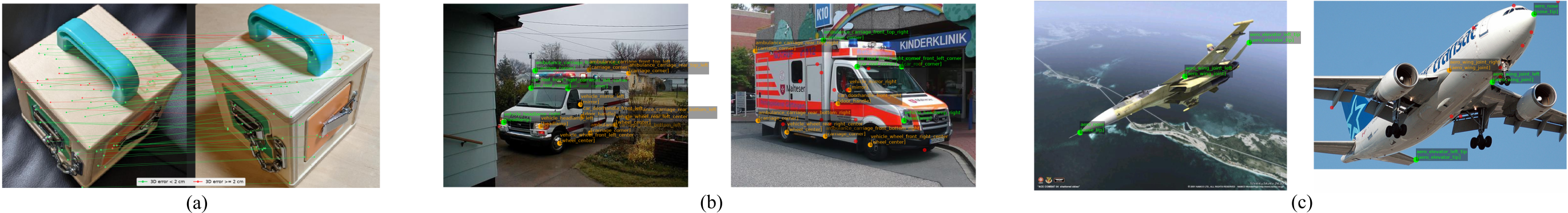}
    \caption{\textbf{Visualization of geometric and semantic correspondence}.
    (a) NAVI large-viewpoint same-instance matching;
    (b) SOCO SOC; 
    (c) SOCO cross-SOC.
    On the left is the source image and on the right is the target image. In figure (b) and (c), the semantic GT, concept-only GT and predicted points are labeled in green, orange and red, respectively.}
    \label{fig:latent_correspondence}
\end{figure*}

\begin{table}[t]
\centering
{\footnotesize
\setlength{\tabcolsep}{2pt}
\renewcommand{\arraystretch}{1.08}
\begin{tabular*}{\columnwidth}{@{\extracolsep{\fill}}lccccc@{}}
\toprule
\multirow{2}{*}{Method}
& NAVI & \multicolumn{4}{c}{SOCO} \\
\cmidrule(lr){2-2}\cmidrule(l){3-6}
& Recall@2\,cm$\uparrow$ & CC$\uparrow$ & SOC$\uparrow$ & Cross-SOC$\uparrow$ & SOC-geo$\uparrow$ \\
\midrule
VGGT (frozen)       & 33.2 & 21.7 & 16.2 & 12.5 & 72.3 \\
VGGT + alignment   & 35.5 & 25.0 & 18.4 & 14.2 & 71.8 \\
\midrule
\textbf{Gen3DOrient} & \textbf{38.6} & \textbf{35.8} & \textbf{23.5} & \textbf{20.4} & \textbf{72.9} \\
\bottomrule
\end{tabular*}\par
}
\caption{\textbf{Zero-shot correspondence beyond orientation.}
NAVI reports geometric matching recall at a 2\,cm threshold.
SOCO CC, SOC, and Cross-SOC report PCK@0.1, and SOC-geo reports
candidate-restricted part-identity accuracy. Higher is better for
all metrics. VGGT + alignment uses the learned feature alignment
layer. All methods use frozen representations and the same matching
protocol for each task.}
\label{tab:latent_downstream}
\end{table}

\vspace{-0.5em}
\subsection{Ablations and Analysis}
\vspace{-0.5em}

\label{sec:exp_ablation}
\textbf{Canonical-pose data curation.}
Table~\ref{tab:ablation_curation_init} isolates the effect of VLM-assisted data
curation. Using curated training labels improves Acc30 on all five
continuous-orientation benchmarks and accuracy on Ori\_COCO. These
results support the value of canonical-orientation consistency in the
training data, without requiring a claim that the VLM itself is a
novel methodological component.

\textbf{Initial pose sampling density.}
The feature-matching objective is non-convex, making initialization
important for gradient-based orientation refinement. We vary the
numbers of candidate azimuth, polar, and in-plane angles. As shown in
Table~\ref{tab:ablation_curation_init}, using a denser search improves results on all
six benchmarks. We use the $96\!\times\!32\!\times\!32$ setting in
our main experiments for better accuracy at the expense of time consumption.

\textbf{Visual feature encoder and alignment.}
Table~\ref{tab:ablation_3d_encoder} compares DINOv2 features, VGGT
features, and VGGT features with the learned alignment layer. For ablation studies, we construct a validation set, termed the VisionEncoderSet, comprising 10K 3D objects selected from our curated dataset. The
alignment variant achieves the lowest median error and highest Acc30
on all three evaluation sets in this ablation. The results suggest
that both the feature source and the alignment step influence final
orientation accuracy; they do not, on their own, establish superior
general-purpose correspondence of the underlying features.

\vspace{-0.5em}
\section{Beyond Orientation: 3D-Structured Latents for Correspondence task}
\vspace{-0.5em}

\label{sec:correspondence}

While Gen3DOrient is trained for single-image 3D representation generation
and evaluated primarily on orientation estimation, the resulting features
can also be examined through correspondence tasks. We consider two
complementary questions: whether the generated representation preserves
\emph{geometric consistency} across views of the same instance, and whether
it supports \emph{semantic part correspondence} across distinct instances.
All evaluations use frozen models without correspondence-specific training.

\textbf{Multi-View Geometric Correspondence.}
We evaluate geometric correspondence on NAVI~\cite{jampani2023navi}, following the
feature-matching protocol of Probe3D~\cite{el2024probing}. Given two views of the
same object, we match dense features and measure the fraction of matches
whose corresponding 3D points lie within a 2\,cm threshold. We also report
results as a function of the relative viewpoint change, to test consistency
under large rotations. For Gen3DOrient, we extract pose-conditioned feature
maps by rendering its image-generated canonical volume. We report the end-to-end evaluation
using estimated orientations. Both settings are compared with frozen VGGT features and VGGT features with learned alignment head under the same matching protocol.

\textbf{Semantic and Geometry-Aware Part Correspondence.}
We further evaluate on SOCO~\cite{dunkel2026soco}, which distinguishes concept
correspondence (CC), semantic object correspondence (SOC), and cross-category
SOC (Cross-SOC). These settings probe progressively different aspects of
part matching, from shared part concepts to object-relative part identity
and transfer across categories. We place particular emphasis on SOC-geo,
which restricts target candidates to keypoints with the same part concept
and tests whether their geometric identities can be distinguished, such as
front versus rear wheels. For CC, SOC, and Cross-SOC we follow the official
PCK@0.1 evaluation; SOC-geo is reported using its candidate-restricted
accuracy protocol. We apply the same training-free matching rule to frozen
baseline features and to rendered Gen3DOrient features. As an additional
geometry-aware diagnostic, we consider SPair-Geo-Aware~\cite{jesslen2026geometry}
and report its PCK@0.1 in the supplementary material.

\textbf{Evaluation and Interpretation.}
Table~\ref{tab:latent_downstream} provides a component-level comparison within our VGGT-based pipeline, separating the frozen visual backbone, the feature-aligned encoder, and the full generated-and-rendered representation. Gen3DOrient improves over both encoder variants on all reported metrics, with larger gains on NAVI and SOCO CC, SOC, and Cross-SOC, and a modest gain on SOC-geo. These results suggest that the generated-and-rendered representation supports correspondence beyond orientation estimation. We do not assume that a shared canonical coordinate system alone establishes semantic correspondence between object instances.

\vspace{-0.5em}
\section{Conclusion}
\vspace{-0.5em}
We introduce Gen3DOrient, a generative analysis-by-synthesis framework that infers object orientation by rendering and comparing generated 3D-structured latents. Experiments demonstrate strong zero-shot generalization across absolute orientation, relative rotation, and correspondence tasks. Our results suggest a broader role for generative 3D representations beyond content generation, toward visual reasoning.

\section*{LLM Usage Statement}
\noindent
Claude Code and Codex are used for coding. Gemini and Codex are used for manuscript editing. VLM is used for dataset curation. No LLMs and AI tools are used for research idea.
{
    \small
    \bibliographystyle{ieeenat_fullname}
    \bibliography{main}
}

\end{document}